%% file: eccv2020submission.tex
\documentclass[runningheads]{llncs}
\usepackage{graphicx}
\usepackage{comment}
\usepackage{amsmath,amssymb} 
\usepackage{color}
\usepackage{caption}
\usepackage{graphicx, wrapfig, lipsum}
\usepackage{multirow}
\usepackage{mathtools}
\usepackage{adjustbox}
\usepackage{soul}
\usepackage{booktabs}
\usepackage{pifont}
\usepackage{xcolor,colortbl}
\usepackage{threeparttable}
\makeatletter
\newif\if@restonecol
\makeatother

\usepackage[ruled,vlined]{algorithm2e}
\usepackage{algpseudocode}

\input{math_commands.tex}

\begin{document}
\pagestyle{headings}
\mainmatter
\def\ECCVSubNumber{2565}  

\title{Efficient Bitwidth Search for Practical \\ Mixed Precision Neural Network}

\titlerunning{ECCV-20 submission ID \ECCVSubNumber} 
\authorrunning{ECCV-20 submission ID \ECCVSubNumber} 
\author{Anonymous ECCV submission}
\institute{Paper ID \ECCVSubNumber}

\titlerunning{Efficient Bitwidth Search}
%
\author{Yuhang Li\inst{1,2}\thanks{contact email \email{loafyuhang@gmail.com}.} \and Wei Wang\inst{2}
\and Haoli Bai\inst{3} \and Ruihao Gong\inst{1} \and Xin Dong\inst{4} \and Fengwei Yu\inst{1}}
\authorrunning{Li et al.}
%
\institute{SenseTime Research, Beijing, China \and National University of Singapore, Singapore \and
The Chinese University of Hong Kong, Hong Kong, China
\\ \and Harvard University, Cambridge MA, USA}
\maketitle

\begin{abstract}
Network quantization has rapidly become one of the most widely used methods to compress and accelerate deep neural networks. Recent efforts propose to quantize weights and activations from different layers with different precision to improve the overall performance.
However, it is challenging to find the optimal bitwidth (i.e., precision) for weights and activations of each layer efficiently. Meanwhile, it is yet unclear how to perform convolution for weights and activations of different precision efficiently on generic hardware platforms. To resolve these two issues, in this paper, we first propose an Efficient Bitwidth Search (EBS) algorithm, which reuses the meta weights for different quantization bitwidth and thus the strength for each candidate precision can be optimized directly w.r.t the objective without superfluous copies, reducing both the memory and computational cost significantly.
Second, we propose a binary decomposition algorithm that converts weights and activations of different precision into binary matrices to make the mixed precision convolution efficient and practical. Experiment results on CIFAR10 and ImageNet datasets demonstrate our mixed precision QNN outperforms the handcrafted uniform bitwidth counterparts and other mixed precision techniques. 


\end{abstract}

\section{Introduction}

\label{sec:introduction}
Due to the outstanding performance of deep neural networks (DNNs) in various applications, there is a surging demand for deploying DNN on edge/mobile devices.
However, mobile devices typically have limited memory and energy, which requires DNNs to be compact and energy-efficient.
There has been a large amount of research on compressing and accelerating neural networks, such as network pruning~\cite{he2017channelprune,zhuang2018discriminationaware}, matrix decomposition~\cite{liu2015sparsedecomp} and quantization~\cite{hubara2017qnn,jacob2018quantizationint}. Especially, network quantization is effective and feasible for deployment and has been widely studied in recent literature.

Most existing quantization methods~\cite{jung2019qil,li2016twn,zhou2016dorefa} adopt uniform precision quantization, i.e., a global precision\footnote{In this work, we use precision and bitwidth in a mixed way.} is used for weights and activations of all layers in CNNs. 
Recently, there is a trend of applying mixed precision quantization~\cite{haq,elthakeb2018releq,wu2019mixed}, i.e., assigning different bitwidths for the weights and activations across different layers. Mixed precision quantization is more flexible and has the potential to save more memory and computational cost without sacrificing the network's expressiveness.
There are two key problems (or challenges) to be resolved in mixed precision quantization as discussed below. 

The first is \textit{how to obtain the optimal bitwidth for each layer effectively and efficiently?} 
Existing techniques for mixed precision can be divided into rule-based methods and learning-based methods. Rule-based methods 
utilize a specific metric such as Hessian spectrum in~\cite{dong2019hawq} to determine the optimal bitwidth for each layer. However, these metrics rely on the heuristics provided by domain experts. 
Recently, inspired by Neural Architecture Search (NAS) 
~\cite{pham2018ENAS,zoph2018NASnet,real2019AmoebaNet,liu2018darts,cai2018proxylessnas}, learning-based approaches~\cite{haq,elthakeb2018releq,wu2019mixed} have been proposed for searching the bitwidths. 
Their methods are built upon either Deep Reinforcement Learning~(DRL) or gradient based approaches.
Despite their success, low efficiency and computational burden are the major drawbacks of these approaches. For instance, Hardware-aware Automated Quantization~(HAQ) adopts DRL to search different configurations (i.e. quantization strategy), but each configuration retrains and evaluates a new model, which is time-consuming~\cite{haq}. Recently proposed differentiable neural architecture search~(DNAS)~\cite{wu2019mixed} follows the gradient-based approach to search for bitwidth configurations. However, the huge super net in DNAS still incurs heavy memory and computational cost for training~\cite{wu2019mixed}.
To this end, we need an efficient way to determine the layerwise precision (i.e., bitwidth). 




\begin{figure}[t]
    \centering
    \includegraphics[width=0.9\textwidth]{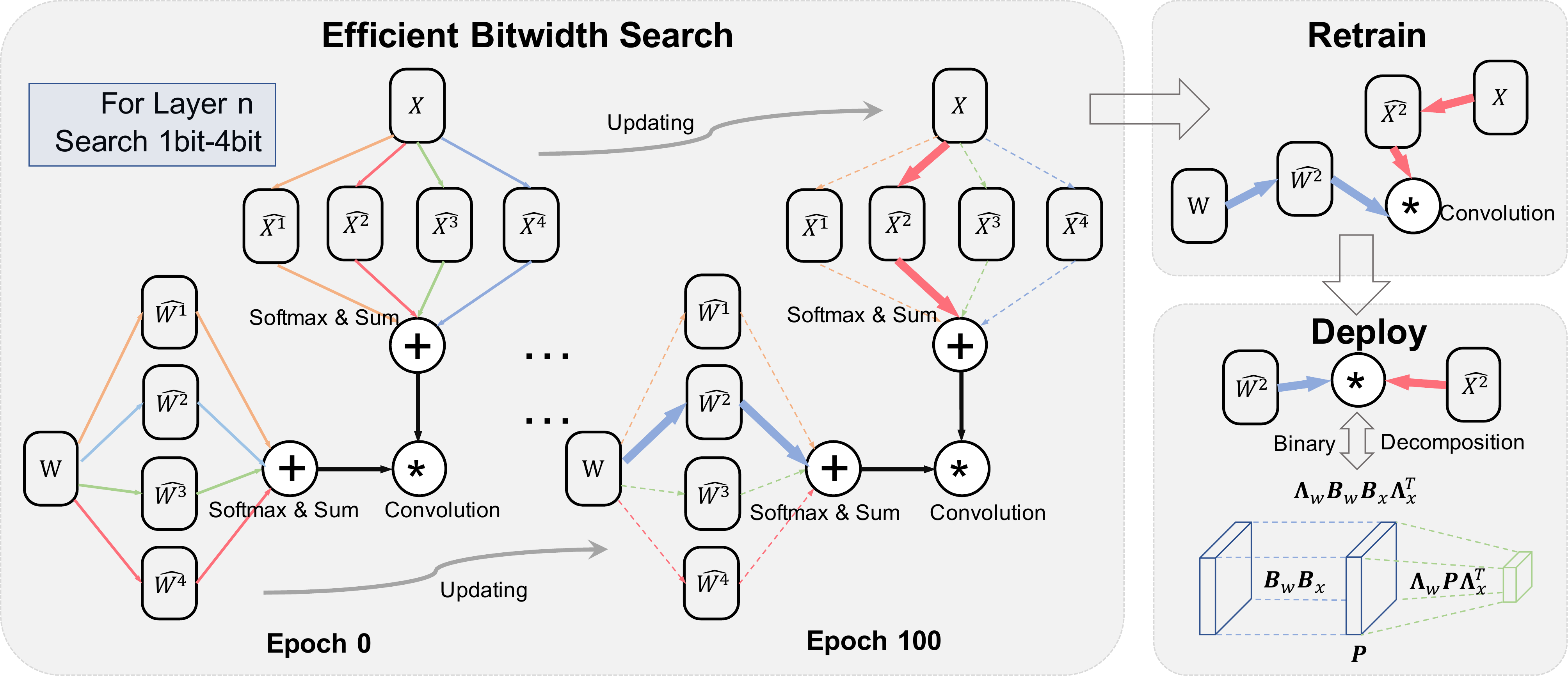}
    \caption{The overall system pipeline. \textbf{In the search stage}, we only maintain one meta weight tensor and quantize it to different bitwidth, then we use softmax to relax the discrete selection of quantized weights. After epochs of training, the strength of each candidate precision is discriminative due to the gradient-based optimization. \textbf{In the retraining stage}, we only select the precision with largest strength and retrain the quantized network. \textbf{In the deployment stage}, Binary Decomposition is applied to support the mixed precision convolution. 
    }
    \label{fig:overview}
\end{figure}

The second problem is \textit{how to do convolution over weights and activations of mixed precision?} While \cite{haq} uses BISMO~\cite{umuroglu2018bismo} and BitFusion~\cite{sharma2018bitfusion} to support mixed precision computation, these platforms are specially designed. General platforms (like ARM CPU, GPU) do not welcome mixed precision computation as they only support INT8, INT4 instructions and binary operations in QNNs. 
Furthermore, they can only support weights and activations with the same precision.
Considering that convolution is essential for CNN, it is necessary to have an efficient convolution implementation between $M$-bit quantized weights and $K$-bit quantized activations, where $M$ and $K$ are the optimized bitwidths.

In this paper, we propose two techniques to address the challenges mentioned above respectively. First, we propose Efficient Bitwidth Search, which is applied in the search process. To make the gradient-based bitwidth search algorithm~\cite{wu2019mixed} efficient, we consider to jointly reduce the memory and computational cost. On the one hand, similar to the spirit of weight sharing in~\cite{pham2018efficient,liu2018darts}, we maintain only one meta weights tensor that can adapt to branches of different quantization precision, which significantly reduces the  memory from $\mathcal{O}(N)$ to $\mathcal{O}(1)$, where $N$ is the number of candidate bitwidths. On the other hand, instead of performing the convolution for each pair of weights precision and activations precision~\cite{wu2019mixed}, we sum up the quantized weights (and activations) from all branches with Softmax weights, and then perform only one convolution, which theoretically reduces the computation from $\mathcal{O}(N^2)$ to $\mathcal{O}(1)$. Moreover, we show that with the weighted sum of quantized weights (and activations), the expressiveness of the quantized network gets significantly improved. Therefore, EBS explores a wider range of quantized space during the search thanks to the dynamic and flexible quantization function.




Second, we propose a Binary Decomposition (BD) algorithm which provides a general computation pattern for mixed precision data in the deployment stage. BD converts the multi-bit weights and activation tensors into binary matrices; then the convolution can be conducted over the binary matrices efficiently. 
Consequently, we can do convolution over mixed precision weights and activations on general-purpose computing platforms without special hardware support. The overall system pipeline is illustrated in Fig.1, which consists of three stages, namely efficient bitwidth search, model retraining and deployment (using BD for convolution).

\noindent\textbf{Contributions}:
\begin{enumerate}
    \item We propose an efficient gradient-based search algorithm to find optimal layerwise bitwidth for mixed precision quantization. The algorithm reduces both the memory and computational cost significantly and is applicable to both weights and activations.
    \item We propose a binary decomposition approach to support efficient convolution over mixed precision weights and activations on generic hardware.
    \item Extensive experiments are conducted on CIFAR10 and ImageNet dataset. Our model has better accuracy-latency trade-off than uniform precision QNNs and other mixed precision QNNs.
\end{enumerate}

\section{Related Works}
\textbf{Quantization } 
Compression via reducing the bitwidth of the parameters in the neural network could significantly reduce the memory cost and accelerate the inference. \cite{li2016twn,rastegari2016xnor,zhu2016ttq} restrict the weight to the binary (1 bit) or ternary (2-bit), however, they have a relatively large accuracy gap with the full precision model. \cite{jung2019qil,zhang2018lqnet} optimize the quantization parameters (levels, intervals) through gradient descent and bridge the gap with full precision models. Yet these methods all use the hand-crafted uniform bitwidth for all layers in the networks. Intuitively, different layers show various sensitivity to quantization, and mixed quantization is recently proposed to assign different layer-wise precision accordingly.
To search for the layerwise precision, 
HAWQ~\cite{dong2019hawq} employs a rule-based method based on the Hessian spectrum of each layer to select the optimal bitwidth, which usually relies on domain expertise.
More recent progress arises from approaches in neural architecture search.

\noindent\textbf{Neural Architecture Search }
Recently, Zoph \textit{et al}. \cite{zoph2016NAS} use Deep Reinforcement Learning (DRL) for NAS, which inspires a series of works for searching operators and connections of the architecture~\cite{pham2018efficient,tan2019mnasnet}.
Similar DRL techniques are applied in network quantization as well. Hardware-aware Automated Quantization (HAQ)~\cite{wang2019haq} adopts DRL for learning layer-wise bitwidth allocation, yet for each bitwidth configuration a new quantized network is retrained, which is time-consuming.
Gradient-based optimization is another promising branch for neural architecture search~\cite{liu2018darts,cai2018proxylessnas}, where both weights and architectures jointly learned by gradient descent. DNAS~\cite{wu2019mixed} is the first to use such differentiable methods to search the quantization precision,
and is the most related work to ours. Comparing to DNAS, our method adopts one meta weights for different quantization branches, and perform only one convolution after the weighted sum of quantized weights and activations with different bitwidth.
As a result, our method improves the memory cost of DNAS from $O(N)$ to $O(1)$, and computational cost from $O(N^2)$ to $O(1)$ significantly, where $N$ is the number of all candidate bitwidths.



\section{Preliminaries}
\newcommand{\bW}{\mathbf{W}}
\newcommand{\bX}{\mathbf{X}}
In this paper, we consider 2D convolution, in which the weights  (a.k.a. kernels) tensor of a convolution layer has four dimensions, namely, the input channel, output channel, height and width respectively, denoted as $\mathbf{W}\in\R^{c_{i}\cdot c_{o}\cdot k\cdot k}$. Similar to DoReFa~\cite{zhou2016dorefa}, we denote the quantization function of weights as $Q(\bW)={\hat{\bW}}$ where ${\hat{\bW}}$ is computed by normalizing the full precision weights into $[-1, 1]$, and then rounding the results to the nearest quantization levels. 
Activations (denoted by $\bX$) of a layer are rectified by ReLU and therefore are non-negative. During quantization, they are clipped to $[0, \alpha]$, normalized to $[0, 1]$, and then rounded to the nearest quantization levels.
This process is formulated as follows:
\begin{subequations}
    \label{eq: quantize}
    \begin{align}
        &\hat{\bW}=2\times\text{quantize}_b(\frac{\tanh(\bW)}{2\max(|\tanh(\bW)|)}+\frac{1}{2})-1,\label{eq:1}\\
        &\hat{\bX}=\alpha\times\text{quantize}_b(\text{clip}(\bX, 0, \alpha)/\alpha),\label{eq:2}\\
        &\text{quantize}_b(x)= \frac{1}{2^{b}-1}\times\text{round}((2^b-1)\times
        x),\label{eq:3}
    \end{align}
\end{subequations} 
where $\alpha$ is the learnable clipping parameter for activations and $b$ is the bitwidth of each weight (or activation) number after quantization. 
Note that  $quantize_b(\cdot)$ includes the de-quantize process which means integers are scaled by $1/(2^b-1)$.
Eq.~\ref{eq:1}-\ref{eq:3} are element-wise operations, where $max(\cdot)$ in Eq.~\ref{eq:1} returns the max absolute weight in $\bW$. $round(\cdot)$ maps the value to the nearest integer with round half up. 
We can see that the whole quantization scheme is parameterized by the bitwidth $b$ and the clipping parameter $\alpha$. In this paper, we focus on optimizing $b$. 

Note that quantization process returns unsigned fixed-point numbers, which make the inference easy to speedup via hardware accelerators. Assume $\hat{\mathbf{w}}$ is a vector of $M$-bit
fixed-point integers s.t. $\hat{\mathbf{w}}=\sum_{m=0}^{M-1}c_{m}(\hat{\mathbf{w}})2^m$ and $\hat{\mathbf{x}}$ is a vector of K-bit fixed-point integers s.t. $\hat{\mathbf{x}}=\sum_{k=0}^{K-1}c_{k}(\hat{\mathbf{x}})2^k$, where each element of $c_{m}(\hat{\mathbf{w}})$ (or $c_{k}(\hat{\mathbf{x}})$) is either 0 or 1.
According to~\cite{zhou2016dorefa}, the dot product of $\hat{\mathbf{w}}$ and $\hat{\mathbf{x}}$ is
\begin{equation}
    \hat{\mathbf{w }}\cdot \hat{\mathbf{x}} = \sum_{m=0}^{M-1} \sum_{k=0}^{K-1}2^{m+k}\text{bitcount}[\text{AND}(c_{m}(\hat{\mathbf{w}}), c_{k}(\hat{\mathbf{x}}))]
    \label{eq.comp}
\end{equation}

When training QNN, we maintain the original full precision weights, called the meta weight $\bW$. 
The gradient of the meta weights are computed via  Straight-Through Estimator~\cite{bengio2013ste}, defined as follows:
\begin{equation}
    \frac{\partial L}{\partial \bW} = \frac{\partial L}{\partial \hat{\bW}}\frac{\partial \hat{\bW}}{\partial \bW},\ \text{where STE is   }\frac{\partial \text{quantize}_b(x)}{\partial x} = 1_{x\le\alpha}.
\end{equation}
STE returns 1 when $x\le\alpha$ otherwise the gradient will be rectified. With the gradient, we can apply SGD to update the meta weights.


\begin{figure}[t]
     \centering
     \includegraphics[width=0.75\linewidth]{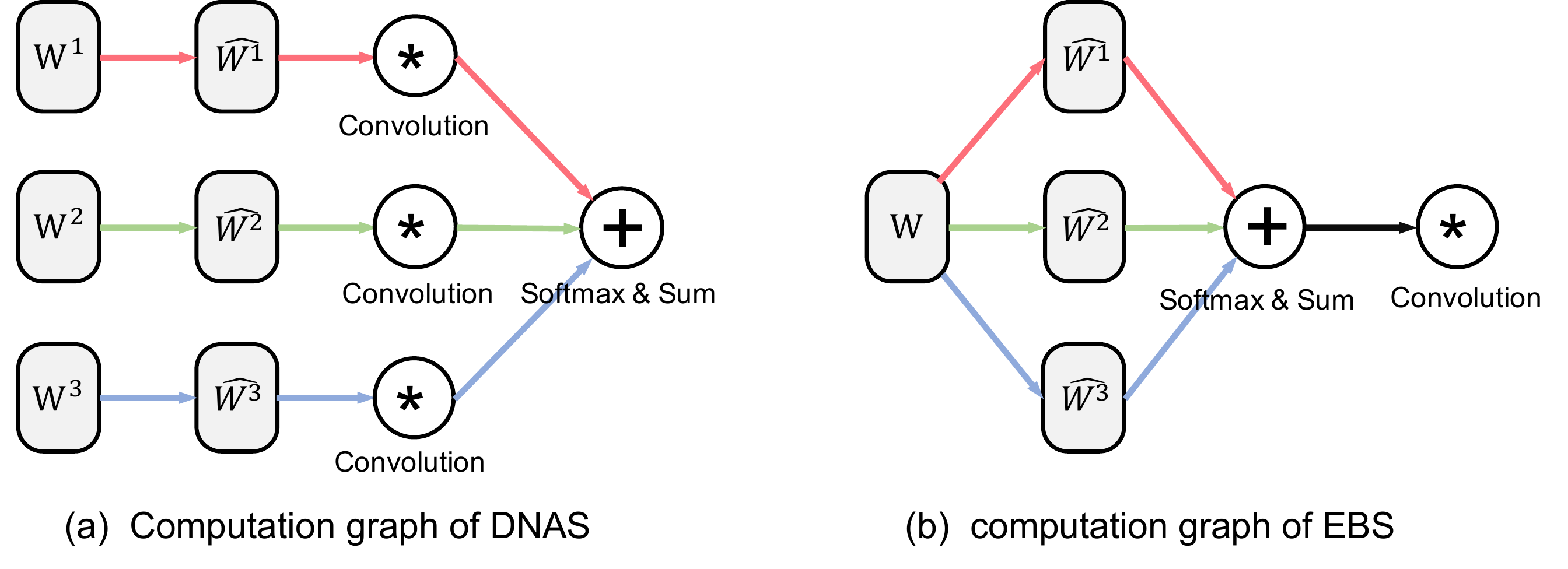}
     \caption{Computation graphs. \textbf{(a)} DARTS and DNAS~\cite{liu2018darts,wu2019mixed} maintain M full precision weight tensors and do M convolution operations if there are M candidate bitwidths; \textbf{(b)}Our method only stores one meta weight tensor; the quantized weights are aggregated; and thus only one convolution operation is done.}
     \label{fig:graph}
     \vspace{-4mm}
\end{figure}

\section{Methodology}
\subsection{Efficient Bitwidth Search (EBS)}
Let $\mathbf{O}$ be the output of the convolution of weights and activations (i.e. $\mathbf{O}=\hat{\bW}*\hat{\bX}$). We use $\hat{\bW}^b$ and $\hat{\bX}^b$ to represent $b$-bit quantized weights and activations. Our task is to find out the best bitwidth configuration for each layer. We discuss quantization of weights and activations respectively in the following paragraphs.

\subsubsection{Weights Quantization}
A simple solution to optimize the bitwidth is to learn a strength (or importance) parameter for each bitwidth, and then select the bitwidth with the largest strength to quantize the matrix. Denote the bitwidth array as $\mathcal{B}=[1, 2,\dots, b_{max}]$, and the strength array as $\mathbf{r}$, the optimal bitwidth is 
\begin{equation}
     b^* = \mathcal{B}[\argmax(\mathbf{r})]\label{eq:opt}.
\end{equation}
Nevertheless, since $\argmax(\cdot)$ is not differentiable, we cannot optimize the strength parameters using gradient-based methods. DARTS~\cite{liu2018darts} and DNAS~\cite{wu2019mixed} resolve this issue by applying Softmax function over the learnable strength parameters $\mathbf{r}$, and then use them to scale the results from each operator correspondingly. In this way, we can compute the gradient of the loss w.r.t $\mathbf{r}$ and apply SGD for optimization.
Armed with the softmax trick, we can perform convolution as follows,
\begin{equation}\label{eq:softmax}
    \mathbf{O} = \sum_{b_i\in\mathcal{B}}\mathbf{O}^i=\sum_{b_i\in\mathcal{B}}\frac{\exp(r_i)}{\sum_{b_j\in\mathcal{B}}\exp(r_j)}(\hat{\bW}^i \ast \hat{\bX}).
\end{equation}
However, as shown in Fig.~\ref{fig:graph}a, DNAS needs to store $N$ copies of meta weights in the super net (one per branch), where $N=|\mathcal{B}|$ is the number of branch candidates in DNAS.
Moreover, if they further quantize activations,
each output $\mathbf{O}_{ij}$ is the feature map of $\hat{\bW}^i$ and $\hat{\bX}^j$, there will be $N^2$ convolution for a single convolutional layer. Therefore Eq.~\ref{eq:softmax} suffers from $\mathcal{O}(N)$ memory cost and $\mathcal{O}(N^2)$ computation cost for each layer.

To prevent increasing GPU memory and computational cost, we only maintain one meta weight tensor as illustrated in Fig. 2b.
In the forward pass, the quantized weight tensors of different precision are scaled and then summed before the convolution,
\begin{equation}
    \mathbf{O} =(\sum_{b_i\in\mathcal{B}}\frac{\exp(r_i)}{\sum_{b_j\in\mathcal{B}}\exp(r_j)}\hat{\bW}^i) \ast \hat{\bX}.
    \label{eq:ours_softmax}
\end{equation}
Consequently, we reduce both the memory and computational cost to $\mathcal{O}(1)$, which significantly improves the search/training efficiency.
Throughout the training process we keep the meta weight as full precision, where the back-propagated gradients adjust itself to pick the most favorable quantization bitwidth. 
At the end of the training, we switch Softmax to max to select the best learned precision (i.e., branch) for the quantization network and remove the branches for other bitwidths. After that, we retrain the network weights to get the final mixed precision QNN. An example of the overall workflow is shown in Fig.~\ref{fig:overview}. The strength for each candidate is distinctive (measured by the line-width), where 2-bit quantization for weights is more favorable by the training objective. 


\begin{figure}[t]
    \centering
    \includegraphics[width=\linewidth]{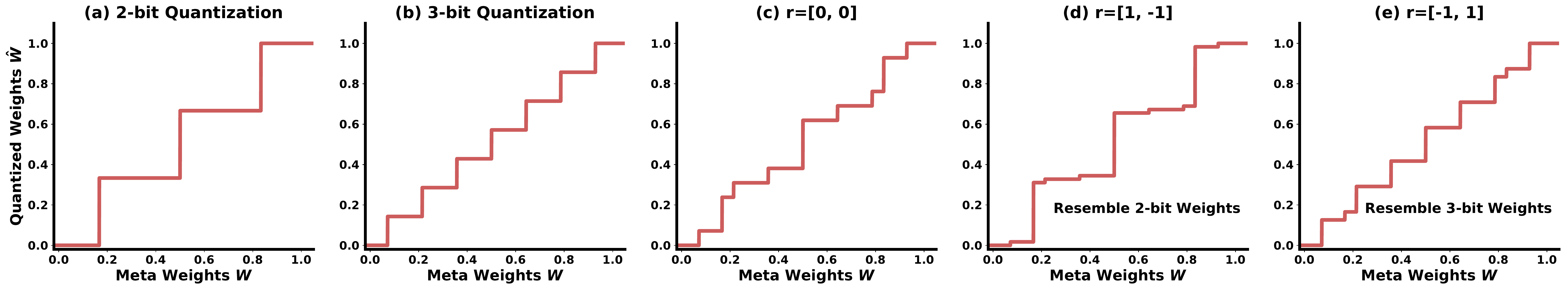}
    \caption{Visualization of the aggregated quantization function. }
    \vspace{-4mm}
    \label{fig:function}
\end{figure}

We need to highlight that, not only the design in Eq.~\ref{eq:ours_softmax} improves the efficiency, but it also brings a more flexible and dynamic quantization function, which shall benefit the search result.
To see this, we visualize the aggregated quantization function of Eq.~\ref{eq:ours_softmax} for our EBS, as shown in Fig.~\ref{fig:function}.
We find that the single precision quantization has the same effect as applying a step function with a uniform step size. Next, we use 2 bitwidths with $\mathcal{B}=\{2, 3\}$. We initialize the strength parameters to $\mathbf{r}=[0,0]$. According to Eq.~\ref{eq:ours_softmax}, the quantized weight is $0.5\hat{\bW}^2+0.5\hat{\bW}^3$, indicating the two quantization results are equally combined and therefore EBS has a larger capacity (i.e., finer precision) to explore the different bitwidths during training. 
As the training continues, the strengths get updated. 
When one strength parameter is much larger than the other one, e.g. $\mathbf{r}=[-1, 1]$, the aggregated quantization result is close to the quantization with the largest bitwidth.
In summary, while EBS seeks to learn a single bitwidth to quantize the weights for inference, it explores multiple bitwidths during training, leading to a dynamic and flexible quantization function.


\subsubsection{Activation Quantization}
We do quantization for the activations in the same way as weights quantization. 
A separate set of strength parameters are learned, denoted as $\mathbf{s}$. And only one convolution in EBS for each convolutional layer is computed during the search.
The convolution of one layer is then formalized as 

\begin{equation}
    \mathbf{O} =(\sum_{b_i\in\mathcal{B}_w}\frac{\exp(r_i)}{\sum_{b_j\in\mathcal{B}_w}\exp(r_j)}\hat{\bW}^i) \ast (\sum_{b_i\in\mathcal{B}_x}\frac{\exp(s_i)}{\sum_{b_j\in\mathcal{B}_x}\exp(s_j)}\hat{\bX}^i).
\end{equation}

\subsubsection{Stochastic Search}
We also introduce a stochastic method to learn optimal precision. First, we denote $f(\mathbf{r})$ as the softmax function. Distinguished from the deterministic search where $\sum[f(\mathbf{r})]=1$ is the coefficient for each candidate precision, we hereby model a categorical distribution where $p_i = f(\mathbf{r})_i$ means the probability of being selected in the forward pass. Then Gumbel-Softmax trick~\cite{maddison2016concrete,jang2016gumbel} is applied to estimate the gradient for sampling from a discrete distribution, given by
\begin{equation}
    \hat{\bW} = \sum_{b_i\in\mathcal{B}}\frac{(\exp(\log p_i + g_i)/\tau)}{\sum_{b_j\in\mathcal{B}}(\exp(\log p_j+g_j)/\tau)}\hat{\bW}^i,\text{ where  }g_i\sim\textit{Gumbel}(0,1).
\end{equation}\label{eq:gumbel}
Note that stochastic search is also applicable to activations. $\tau$, the temperature, controls the tightness in the sampling process. In experiments, we shall compare the differences of the two approaches. 

\subsection{Optimization}
During training, we alternatively optimize weights and architecture parameters (i.e., the bitwidths), leading to a bilevel optimization problem. As aforementioned, bitwidth is closely related to the hardware performances of QNNs, such as model size and computational cost (FLOPs). 
Therefore, we add the computational cost into the loss function when optimizing the bitwidths. We first set a hyperparameter FLOPs$_{target}$ for the desired computation cost in the target mixed precision QNN. The bilevel optimization is therefore given by:
\begin{align}
     \min_{\mathbf{r}, \mathbf{s}} & \ \mathcal{L}_{valid}(\mathcal{W}^*, \mathbf{r}, \mathbf{s}) + \lambda \max(0, \text{FLOPs}-\text{FLOPs}_{target}), \label{eq:loss1}\\
     \text{s.t. } & \mathcal{W}^* = \argmin_{\mathcal{W}} \mathcal{L}_{train} (\mathcal{W}, \mathbf{r}, \mathbf{s}),
\end{align}
where $\mathcal{W}$ denotes the weights in the whole network. We abuse $\mathbf{r}$ and $\mathbf{s}$ to denote the strength parameters for all convolution layers. The detailed algorithm is shown in Alg.~1. 
We should point out that FLOPs$_{target}$ cannot ensure the final QNN has exactly the same FLOPs as expected, because during inference $\max(\mathbf{r})$ will result in a single precision weights while the FLOPs is computed based on the average as shown below.
For FLOPs calculation, we use the expectation of the computational cost of all branches. We define a function FLOP$(M,K)$ that returns the operation count of the convolution between $M$-bit weights and $K$-bit activations. From Eq.~\ref{eq.comp}, we can see that the operation count is differentiable w.r.t M and K. We compute the expected FLOPs for both deterministic and stochastic search by
\begin{equation}
\label{eq:flops}
    \mathbb{E}(\text{FLOPs}) = \text{FLOP} (\sum_{b_i\in\mathcal{B}_w}(f(\mathbf{r})_i\times b_i), \sum_{b_j\in\mathcal{B}_x}(f(\mathbf{s})_j\times b_j)). 
\end{equation}


\begin{algorithm}[t]
\caption{Search Algorithm.}
\label{alg:1}
\KwIn{Initial strength $\mathbf{r=0}$, target FLOPS; Total training epoch T ; training set and validation set.}
\For{all $i=1,2,\dots, T$-epoch}
      {
        \For{all $j=1,2,\dots, T^{\prime}$-batch iteration}
        {
            Update $\mathcal{W}$ to minimize $\mathcal{L}_{train}$ over the training set\;
            Update $\mathbf{r}$, $\mathbf{s}$ to minimize Eq.~\ref{eq:loss1} over the validation set\;
        }
      }
      \textbf{return} optimized bitwidths according to Eq.~\ref{eq:opt} for each convolution layer\;
\end{algorithm}

\subsection{Binary Decomposition} 
Our objective is to deploy the model for efficient and accurate inference. 
However, during inference, it is non-trivial to conduct efficient convolution between weights and activations of different precision on generic hardware. Note that the bitwidth for the weights and activations for the same layer could be different. 

Img2col is a popular way to implement convolution that converts the weights tensor and activations tensor into matrices and then the convolution is done via matrix multiplication. We adopt img2col in this paper. For example, suppose we have a quantized weights matrix $\hat{\bW}\in S^{2\times 3}$ and activations matrix $\hat{\bX}\in S^{3\times 2}$. We also assume that the bitwidths for the weights and activations are 2 and 3 respectively.
So that $S=\{0,1,2,3\}$. 
In Sec.~3 (Eq.~\ref{eq.comp}, we show that the quantized values using fixed-point number format can be rewritten by $\hat{\bW}=\sum_{m=0}^1c_m(\bW)2^m$ where $c_m$ return the binary value $\{0,1\}$. 
Hence, we use this expansion to decompose the weights and activations:
\begin{equation}
    \hat{\bW} = 
    \begin{bmatrix}w_{11} & w_{12} & w_{13}\\w_{21} & w_{22} & w_{23}
    \end{bmatrix}
    = \begin{bmatrix} 2^0&2^1&0&0\\ 0&0&2^0&2^1
    \end{bmatrix}
    \begin{bmatrix} c_0(w_{11}) &c_0(w_{12})& c_0(w_{13})\\c_1(w_{11}) &c_1(w_{12})& c_1(w_{13})\\c_0(w_{21}) &c_0(w_{22})& c_0(w_{23})\\c_1(w_{21}) &c_1(w_{22})& c_1(w_{23})
    \end{bmatrix}
    = \Lambda_{w}\mathbf{B}_{w},
\end{equation}
where $\mathbf{B}_{w}\in \{0, 1\}^{4\times3}$ is the decomposed binary matrix and $\Lambda_w$ is the coefficient matrix for the binary values. The same decomposition can be applied to activations by $\hat{\bX}=\mathbf{B}_x\Lambda_x^T$. After BD, the feature maps is computed by 
\begin{align}
    \mathbf{O} & = \Lambda_w\mathbf{P}\Lambda_x^T \text{ where } \mathbf{P}=\mathbf{B}_w\mathbf{B}_x\in\R^{4\times4}, \label{eq:binconv}\\
    o_{(i, j)}& =\sum_{m=0}^1\sum_{k=0}^1 p_{(2i+m-1, 2j+k-1)}2^{m+k}.\label{eq:2mk}
\end{align}
We notice that the core computation
is computed by binary operation. 
\begin{wrapfigure}{r}{4.5cm}
\vspace{-4mm}
\includegraphics[width=4.5cm]{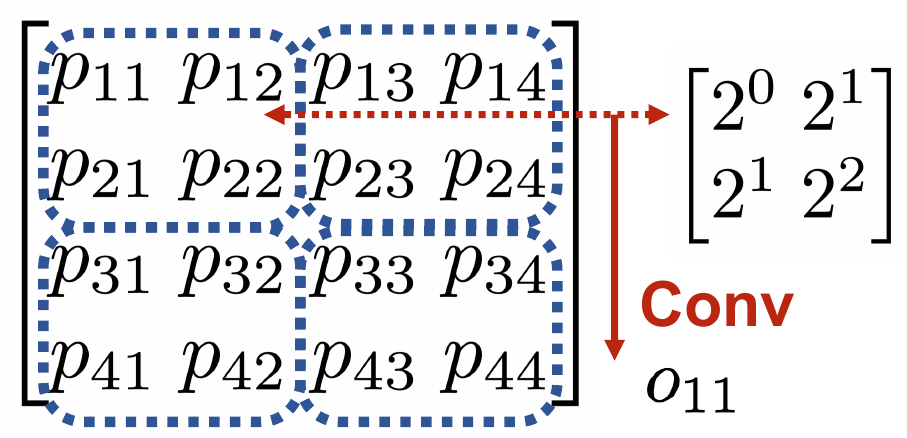}
\caption{Computation of Eq.\ref{eq:2mk}}\label{wrap-fig:1}
\vspace{-4mm}
\end{wrapfigure}
Next we introduce an efficient implementation of $\Lambda_w\mathbf{P}\Lambda_x^T$. 
Eq.~\ref{eq:2mk} explicitly gives the outcome of $\Lambda_w\mathbf{P}\Lambda_x^T$. 
Here, $i$ and $j$ are the row and column index of matrix $\mathcal{O}$ respectively. 
We visualize the computation for $O$ in the right figure. It can be seen that  $\mathbf{P}$ is divided into 4 parts, each of which does a vector dot product\footnote{We flatten the  matrices into vectors first.}  with a $2\times 2$ matrix consisting of powers-of-2 values.  
This means we can use a \textbf{depthwise convolution} with stride 2 and powers-of-2 kernels to implement $\Lambda_w\mathbf{P}\Lambda_x^T$. 


We next present the formal definition of Binary Decomposition. BD uses 2 convolutions to obtain the final outcome of the mixed precision convolutions, the first is the normal convolution with binary weights and activations,
then a depthwise convolution is applied to compute $\mathcal{O}$.
Suppose weights and activations are quantized $M$-bit and $K$-bit, respectively, where $\hat{\bW}\in S^{c_o \times s}$ and $\hat{\bX}\in S^{s \times n}$.
After decomposition, the coefficient matrix can be represented by:
\begin{equation}
    \Lambda_w = \text{diag}\overbrace{(\delta_w, \dots, \delta_w)}^{c_o}, \text{ where } \delta_w=[2^0, 2^1, \dots, 2^{M-1}].
\end{equation}
Therefore $\Lambda_w$ has shape of ($c_o\times c_oM$) and $\Lambda_x^T$ has shape of ($nK\times n$). 
The binary matrices $\mathbf{B}_w\in\{0,1\}^{c_oM\times s}$ and $\mathbf{B}_x\in\{0,1\}^{s\times nK}$ can be derived by taking the expansion of fixed-point values. In the first convolution,
only AND and bitcount are required to get the intermediate result $\mathbf{P}$. Then we construct a second layer, with only one depthwise kernel computed by $\delta_w^T\delta_x$. The kernel has a shape of $(M, K)$ with stride of $(M, K)$. Kernels are comprised of powers-of-2 values, therefore the convolution is equal to shift the values in $\mathbf{P}$ and can be efficient to implement. As a result, BD circumvents the direct mixed precision computation by decoupling the binary operations and the depthwise convolution.

\subsubsection{Complexities Analysis}
In this section, we give the computation and memory complexities analysis w.r.t. the BD algorithm and we show that BD actually only incur negligible memory overhead. For a layer-wise clipping parameter $\alpha$ in activations, we do not modify any memory cost nor the computation since it can be merged in Batch Normalization~\cite{ioffe2015bn} and therefore we omit its analysis. 

Let us take a look at memory first, the shape of the quantized weight matrix is $(c_o\times s)$ after img2col, where $s=c_ik^2$. Before BD, the storage cost is $sc_oM$ bits. After decomposition, weights are represented by $\Lambda_w$ and $\mathbf{B}_w$. $\mathbf{B}_w$ has $s\times c_oM$ binary (1 bit) weights; therefore it incurs the same memory cost as the  weights before BD. The additional memory cost comes from $\Lambda_w$; during inference, we do not store the sparse matrix, instead according to Eq.~\ref{eq:2mk}, we only need to store $MK$ power-of-2 values. Considering that M and K are small, e.g., smaller than 5 in our experiments, the additional memory cost is negligible. 
where in deployment it represents the kernel of the second convolutional layer. The kernel is determined by $\delta_w^T\delta_x$ which only needs $MK$ powers-of-2 values. As a result, the only additional memory requirement is the $MK$ fixed-point numbers and can be negligible compared to weights (e.g., most layer in ResNet-34 has the weight shape of $256\times256\times3\times3$).

For computation, $\hat{\bX}$ has the dimension of $(s\times n)$ where $n$ denotes the numbers of element in a single channel of feature maps. Then, the total FLOPs based on the Eq.~\ref{eq.comp} is $snc_oMK$ AND operations, $nc_oMK$ bitcount and shift-adds operations. When BD is applied to the QNN, the first convolution outputs $\mathbf{P}\in\R^{c_oM\times nk}$, therefore, we can compute the FLOPs is also $snc_oMK$ AND operations and $nc_oMK$ bitcount operations. No shift-adds in the first stage because weights and activations are binarized. In the second stage, each element in $\mathbf{P}$ needs one time shift-add. Therefore, we can conclude that no extra computation cost is introduced in BD. 


\section{Experiments}
In this section, we evaluate our proposed EBS for the modern popular neural network ResNets~\cite{he2016resnet} on ImageNet ILSVRC-2012 dataset~\cite{russakovsky2015imagenet} and CIFAR10~\cite{krizhevsky2009cifar} to demonstrate the effectiveness of our algorithm.

\newcommand{\flexible}{\textit{flexible}}
\vspace{2mm}
\noindent\textbf{Implementation }
Based on prior works and our preliminary experiment results, 5-bit quantization can generally preserve the full precision accuracy of ResNets. Therefore our search space is set to $\mathcal{B}=\{1,2,3,4,5\}$. We provide the details of the evaluation and search implementation as well as code in the Supplemental materials.

\newcommand{\pink}{\cellcolor[rgb]{1, 0.87, 0.87}}
\newcommand{\purple}{\cellcolor[rgb]{0.87, 0.87, 1}}
\newcommand{\blue}{\cellcolor[rgb]{0.933, 0.87, 0.933}}

\subsection{CIFAR10}
In this section, we compare the accuracy and FLOPs of inference in ResNet-20, 32, and 50. 
The compared methods include: 1)uniform precision QNNs, which have a pre-defined bitwidth for all weights and activations, 2) EBS-Det: EBS with deterministic search, 3) EBS-Sto: EBS with stochastic search and 4) random search: sample a random precision QNN within the target FLOPs range. We run the last 3 algorithms with 3 different FLOPs targets.

\begin{figure}[t]
    \centering
    \includegraphics[width=\linewidth]{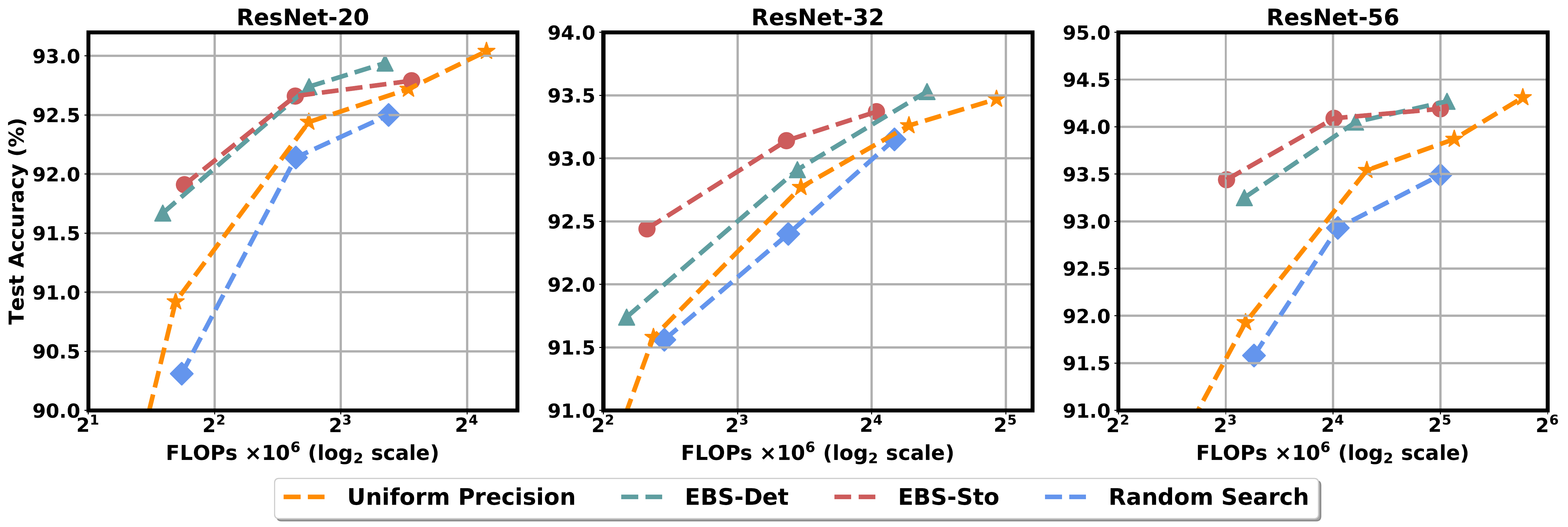}
    \caption{Accuracy-FLOPs curve of ResNets on CIFAR10.}
    \label{fig:cifar}
    \vspace{-5mm}
\end{figure}

Table~\ref{tab1} and Fig.~\ref{fig:cifar} summarize the results of these 4 approaches. First, we compare our EBS-Det with uniform precision QNN. It can be seen that 2-bit quantized models have severe accuracy degradation (2.04\%), while EBS-Det with similar computation cost can achieve 0.75\% accuracy improvement. This means that 2 bit quantization is not enough for some weights and activations to extract or preserve sufficient information, and a reasonable precision allocation is important. Furthermore, EBS-Det can reach the full precision model accuracy with 4x speedup, which demonstrates it is not necessary to allocate 5-bit quantization to all weights and activations. In ResNet-32 and 56, the same trend is also observed. In particular, EBS-Det with 6.79x speedup can surpass the accuracy of 4-bit quantized model, which is a significant improvement.

\begin{table}[t]
\centering
\caption{Accuracy and computational cost comparison over CIFAR10.
}
\begin{adjustbox}{max width=0.93\linewidth}
    \begin{tabular}{lc ccc ccc ccc}
    \toprule
    \multirow{2}{6em}{\textbf{Methods}} & \multirow{2}{5em}{\textbf{Precision}} &  \multicolumn{3}{c}{\textbf{ResNet-20}} &\multicolumn{3}{c}{\textbf{ResNet-32}} & \multicolumn{3}{c}{\textbf{ResNet-56}}\\
    \cmidrule(l{2pt}r{2pt}){3-5}
    \cmidrule(l{2pt}r{2pt}){6-8}
    \cmidrule(l{2pt}r{2pt}){9-11}
    & & Accuracy& FLOPs & Saving & Accuracy & FLOPs & Saving & Accuracy & FLOPs & Saving\\
    \midrule
    Full Prec. & 32-bit & \pink92.96 & \pink40.81 M & \pink1.0$\times$ & \blue93.52 & \blue69.12 M & \blue1.0$\times$ & \purple94.46 & \purple125.7 M & \purple1.0$\times$ \\
    \midrule 
    & 5 bits & 93.04 & 17.8 M& 2.29$\times$ & 93.47 & 30.5 M& 2.27$\times$ &  94.31 & 54.5 M& 2.30$\times$ \\
    Uniform& 4 bits & 92.72 & 11.6 M& 3.53$\times$ & 93.26 & 19.4 M& 3.56$\times$&  93.87 & 35.0 M& 3.59$\times$ \\
    Precision& 3 bits & 92.44 & 6.71 M& 6.08$\times$ & 92.77 & 11.1 M & 6.23$\times$ & 93.54 & 19.9 M& 6.31$\times$ \\
    QNN& 2 bits & 90.92 & 3.23 M& 12.6$\times$ & 91.58 & 5.18 M& 13.3$\times$ & 91.93 & 9.09 M& 13.8$\times$ \\
    & 1 bit & 84.31 & 1.14 M& 35.8$\times$ & 86.68 & 1.63 M & 42.4$\times$ & 88.14 & 2.60 M& 48.3$\times$ \\
    \midrule 
    
    & \flexible & \pink\textbf{92.94} & \pink\textbf{10.2 M}& \pink\textbf{4.01}$\times$ & \textbf{93.53} & 21.3 M & 3.24$\times$ &  \textbf{94.27} & 33.4 M& 3.76$\times$ \\
    EBS-Det& \flexible & \textbf{92.74} & 6.72 M& 6.07$\times$ & 92.91 & 10.9 M & 6.34$\times$ & 94.05 & 18.5 M& 6.79$\times$ \\
    & \flexible & 91.67 & \textbf{3.01 M}& \textbf{13.6}$\times$ & 91.74 & \textbf{4.51 M} & \textbf{15.3}$\times$ & 93.25 & 9.02 M& 13.9$\times$ \\
    \midrule 
    
    & \flexible & 92.79 & 11.8 M& 3.46$\times$ & 93.37 & \textbf{18.5 M} & \textbf{4.21}$\times$ & 94.19 & \textbf{32.0 M}& \textbf{3.93}$\times$ \\
    EBS-Sto& \flexible & 92.66 & \textbf{6.23 M}& \textbf{6.56}$\times$ & \blue\textbf{93.14} & \blue\textbf{10.3 M} & \blue\textbf{6.71}$\times$  & \purple\textbf{94.09} & \purple\textbf{16.1 M}& \purple\textbf{7.80}$\times$ \\
    & \flexible & \textbf{91.91} & 3.39 M& 12.0$\times$ & \textbf{92.44} & 5.01 M & 13.8$\times$ & \purple\textbf{93.44} & \purple\textbf{8.04 M}&  \purple\textbf{15.6}$\times$ \\
    \midrule 
    
    & \flexible & 92.50 & 10.4 M& 3.92$\times$ & 93.15 & 18.0 M & 3.84$\times$ &  93.49 & 32.0 M& 3.93$\times$ \\
    Random Search& \flexible & 92.14 & 6.25 M& 6.52$\times$ & 92.40 & 10.4 M & 6.64$\times$ & 92.93 & 16.5 M& 7.62$\times$ \\
    & \flexible & 90.31 & 3.34 M& 12.2$\times$ & 91.56 &5.48 M & 12.6$\times$ &  91.58 & 9.60 M& 13.1$\times$ \\
    \bottomrule
    \end{tabular}
    \end{adjustbox}
\label{tab1}
\end{table}

We also compare deterministic search with stochastic search, EBS-Sto slightly outperforms EBS-Det when FLOPs are low. For example, in ResNet-56, EBS-Sto overshadows the deterministic method with 0.19\% accuracy improvement as well as 1 million fewer FLOPs, 
one possible reason is that in the low bit scenario, the optimization falls in local minimum more frequently as it is difficult to optimize, and the stochastic method can somehow jump out from the local minimum. However, for 4-bit or 5-bit, the loss landscape resembles that of the full precision training, therefore, the deterministic search might become more suitable. Nevertheless, the discrepancies between these two methods are not significant. Last but not least important, we compare our search algorithm with random search, which initializes the model with a Gaussian vector of $\mathbf{r}, \mathcal{s}$ and sample the bitwidths to construct QNNs. We only keep QNNs whose FLOPs are in target range. We can see that random searched QNNs have even lower accuracy than the uniform precision QNNs, it could be because some activations are sampled to 1-bit and thus severely damage the network performances as binary activations have the lowest expressiveness.

\subsection{ImageNet}

We evaluate our algorithm on ResNet-18 and ResNet-34~\cite{he2016resnet} for ImageNet dataset
Readers can refer to the Appendix for the results on ResNet-34. We compare our results with existing uniform precision QNN methods, including PACT~\cite{choi2018pact} 
and other strong baselines such as LQ-Net~\cite{zhang2018lqnet} and DSQ~\cite{gong2019dsq}. 
We also compare our EBS algorithm with DNAS~\cite{wu2019mixed}, another differentiable Mixed Precision QNN.

Table~\ref{tab2} summarizes the top-1, top-5 as well as the FLOPs of the baselines. Interestingly, PACT with 2-bit weights and 2-bit activations only achieve 64.4\% top-1 accuracy on ImageNet, while its 1-bit weights and 3-bit activation version improves the accuracy by 0.9\%. This may be counter-intuitive at the first glance, because the latter has fewer FLOPs and smaller model size than the former one; however, \cite{choi2018pact,mishra2018wrpn} motioned that activations quantization is more important in QNNs, therefore increasing the bitwidth of activations can be helpful to QNNs. This phenomenon also points out that uniform precision for weights and activations is not optimal. 
Based on Fig.~\ref{fig:res18}, we can see that EBS's accuracy outperforms other uniform precision techniques, due to the reasonable precision allocation. In the low FLOPs model, EBS-Sto can surpass the state-of-the-art models by 1.8\% for top-1 accuracy while preserving the FLOPs. Our EBS-Det can outstrip 5-bit PACT models with 0.4\% for top-1 accuracy and only degrade 0.1\% accuracy against the full precision models. Last, we compare our results with DNAS, with the label refinery enhancement for both approaches. It can be seen that our methods consistently improve the accuracy of DNAS, this is because our model can efficiently explore more combinations of weights and activations. 

\noindent
\begin{minipage}[t]{\linewidth}
\begin{minipage}[b]{0.56\linewidth}
\centering
\begin{adjustbox}{max width=\linewidth}
    \begin{tabular}{lcc cccc}
    \toprule
    \multirow{2}{6em}{\textbf{Methods}} & \multicolumn{2}{c}{\textbf{Precision}} &  \multicolumn{4}{c}{\textbf{ResNet-18}}\\
    \cmidrule(l{2pt}r{2pt}){2-3} \cmidrule(l{2pt}r{2pt}){4-7} 
    & Weights & Activations & Top-1 & Top-5 & FLOPs & Saving\\
    \midrule
    Full Prec. & 32-bit & 32-bit & 70.4 & 89.6 & 1.82 G & 1.0$\times$ \\
    \midrule 
    PACT & 5-bit & 5-bit & 69.8 & 89.3 & 849 M & 2.14$\times$ \\
    \midrule
    PACT & 4-bit & 4-bit & 69.2 & 89.0 & 586 M & 3.10$\times$ \\
    LQ-Net & 4-bit & 4-bit & 69.3 & 88.8 & 586 M & 3.10$\times$ \\
    DSQ & 4-bit & 4-bit & 69.6 & - & 586 M & 3.10$\times$ \\
    EBS-Det & \flexible & \flexible & \pink\textbf{70.2} & \pink\textbf{89.3} & \pink\textbf{558 M} & \pink\textbf{3.26$\times$} \\
    EBS-Sto & \flexible & \flexible & 70.0 & \textbf{89.3} & 564 M &  3.22$\times$\\
    \midrule
    PACT & 3-bit & 3-bit & 68.1 & 88.2 & 381 M & 4.77$\times$ \\
    LQ-Net & 3-bit & 3-bit & 68.2 & 87.9 & 381 M & 4.77$\times$ \\
    DSQ & 3-bit & 3-bit & 68.7 & - & 381 M & 4.77$\times$ \\
    EBS-Det & \flexible & \flexible & 69.4 & 88.9 & \textbf{369 M} & \textbf{4.93}$\times$ \\
    EBS-Sto & \flexible & \flexible & \textbf{69.5} & \textbf{89.1} & 380 M & {4.78}$\times$ \\
    \midrule
    PACT & 2-bit & 2-bit & 64.4 & 85.6 & 235 M & 7.75$\times$ \\
    PACT & 1-bit & 4-bit & 65.0 & 85.9 & 235 M & 7.75$\times$ \\
    PACT & 1-bit & 3-bit & \purple65.3 & \purple85.9 & \purple206 M & \purple8.83$\times$ \\
    LQ-Net & 2-bit & 2-bit & 64.9 & 85.9 & 235 M & 7.75$\times$ \\
    DSQ & 2-bit & 2-bit & 65.2 & - & 235 M & 7.75$\times$ \\
    EBS-Det & \flexible & \flexible & 66.3 & 86.5 & 216 M & 7.42$\times$ \\
    EBS-Sto & \flexible & \flexible & \purple\textbf{67.0} & \purple\textbf{87.2} & \purple\textbf{211 M} & \purple\textbf{7.91$\times$} \\
    \midrule
    DNAS & \flexible & \flexible & 70.6 & - & 594 M & 3.06$\times$ \\
    (+label refinery) & \flexible & \flexible & 68.7 & - & 406 M & 4.48$\times$ \\
    \midrule
    EBS-Det & \flexible & \flexible & \blue\textbf{71.1} & \blue\textbf{89.7} & \blue558 M & \blue3.26$\times$ \\
    (+label refinery) & \flexible & \flexible & \blue\textbf{70.3} & \blue\textbf{89.3} & \blue369 M & \blue4.93$\times$ \\
    \bottomrule
    \end{tabular}
    \end{adjustbox}
\captionof{table}{Accuracy and computations cost comparison of ResNet-18 between EBS and existing methods in ImageNet.}
\label{tab2}
\end{minipage}
\hfill
\begin{minipage}[b]{0.42\linewidth}
\centering
    \includegraphics[width=\linewidth]{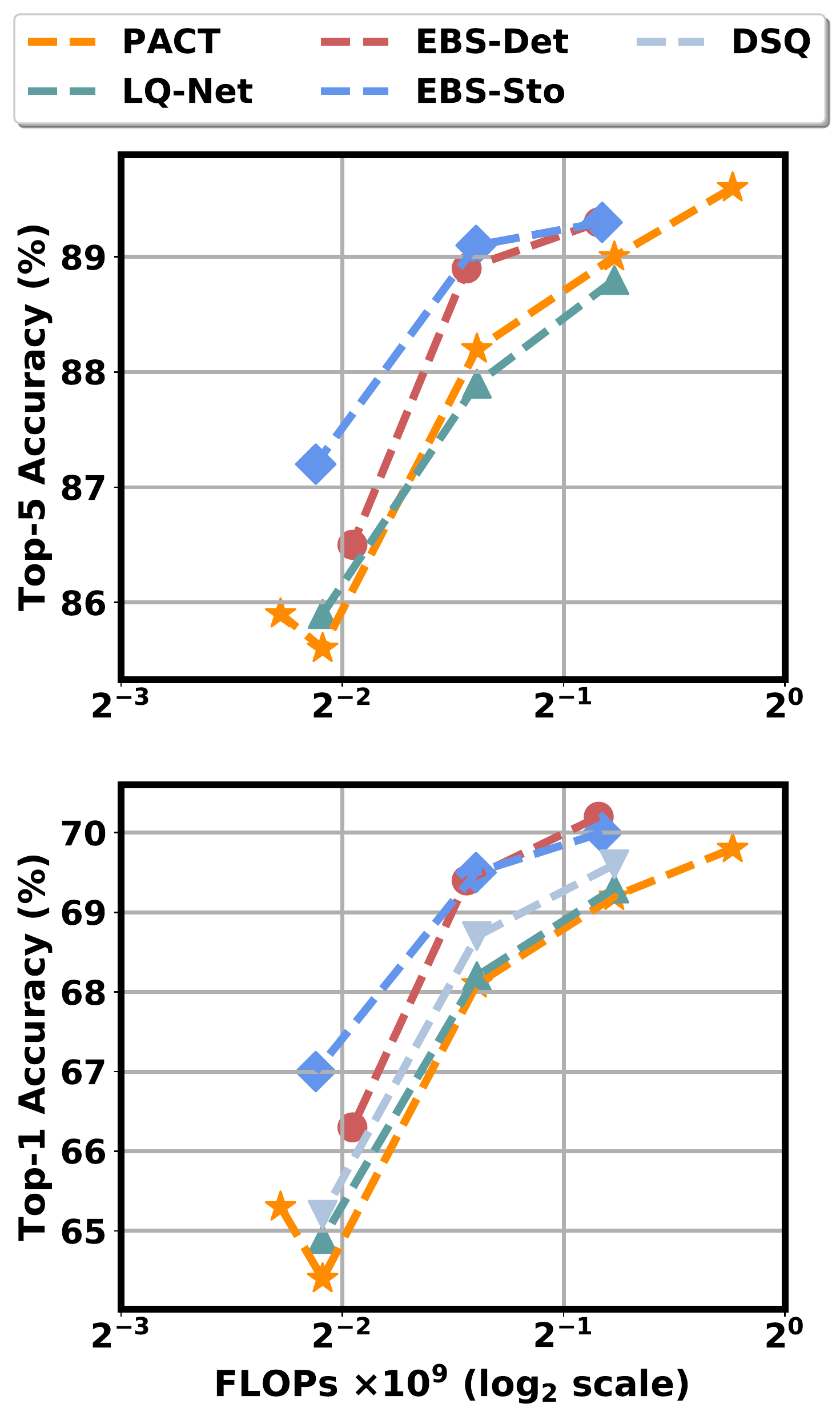}
\captionof{figure}{Accuracy-FLOPs curve of ResNet-18.}
\label{fig:res18}
\end{minipage}
\end{minipage}

The following figure gives the bitwidth distribution of ResNet-18 on ImageNet with the least FLOPs. We can see that most weights of the network are quantized to 1-bit and the average bitwidth of activations is higher than that of the weights. This allocation has also been confirmed by PACT that 1-bit weights and 4-bit activations are better than 2-bit weights and 2-bit activations. However, we can only know that activations need more bits, but how many bits for each layer and which layer should get more bits are challenging problems. EBS-Det and EBS-Sto can discover the precision distribution automatically and adapt to any target FLOPs.

\begin{figure}[t]
\vspace{-2mm}
    \centering
\includegraphics[width=0.8\linewidth]{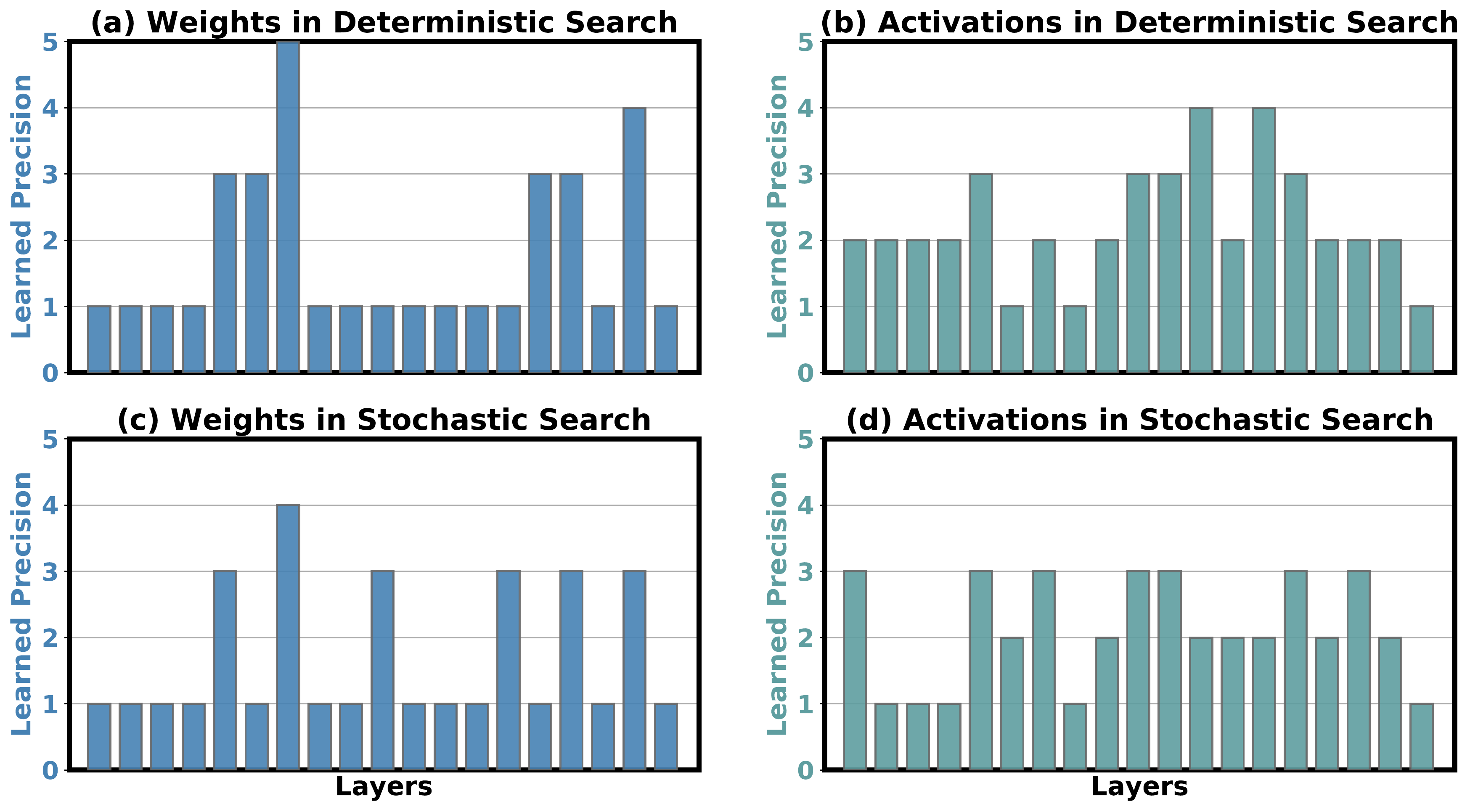}
\captionof{figure}{Precision distribution of ResNet-18 with least FLOPs.}
\label{fig:precision18}
\vspace{-4mm}
\end{figure}

\subsection{Efficiency}
As we stated before, the GPU memory and computation cost are $\mathcal{O}(N)$ and $\mathcal{O}(N^2)$ in DNAS respectively; and both are reduced to $\mathcal{O}(1)$ in EBS. It is necessary to verify the real GPU memory and time savings for DNAS and our EBS. Note that we use fake quantization training on GPU, where full precision with constrained values are used to emulate the quantization. We compare with the single-precision QNN, EBS and DNAS network as shown in Table~\ref{tab:efficiency}. 
Here we let EBS and DNAS search in the same space (5 candidate precision per layer) and we show that EBS can reduce the orders of magnitude of GPU time and memory compared with DNAS. In particular, EBS only increase 3.6GB GPU memory and 1.4 seconds compared with uniform precision QNN when batch size is set to 32. If the batch size continues to 64 or 128, EBS can still search the bitwidth efficiently while DNAS model will become out of memory (OOM).

\noindent
\begin{minipage}{\linewidth}
\centering
\captionof{table}{GPU memory (GB) and time (second) of training  ResNet-18 for 10 iterations.}
\vspace{-2mm}
\begin{adjustbox}{max width=0.9\linewidth}
    \begin{tabular}{l| l| cccc| l| cccc}
    \hline
    \multicolumn{1}{c}{}& & \multicolumn{5}{c}{Batch Size} & \multicolumn{4}{c}{Batch Size}\\\hline
    \multicolumn{1}{l|}{Model}& & 16 & 32 & 64 & 128 & & 16 & 32 & 64 & 128\\
    \hline
    Uniform Precision QNN &Memory & 2.5 & 3.5 & 5.1 & 8.4&Time & 16.7 & 20.9 & 26.7 & 42.0 \\
    EBS & Memory & 4.6 & 7.3 & 12.5 & 22.0 & Time & 17.7& 22.3 &30.7 &  47.1\\
    DNAS & Memory & 36.9 & 71.8 & OOM& OOM& Time &55.5 &100 &- &- \\\hline
    \end{tabular}
    \end{adjustbox}
\label{tab:efficiency}
\end{minipage}

\section{Conclusion}
\vspace{-3mm}
In this paper, we propose a novel and efficient quantization scheme for mixed precision QNNs, which learns layer-wise bitwidths for weight and activation representation. We improve the efficiency of gradient-based search methods by reusing meta weights for different quantization bitwidth, which significantly reduces the memory and computation.
To enable efficient convolution with mixed precision, we propose to decompose the tensors into a binary matrix and a coefficient matrix. Extensive experiments confirm the superiority of our solution in comparison to uniform precision and other mixed precision schemes.

%

\clearpage

%
%

\bibliographystyle{splncs04}
\bibliography{egbib.bib}

\newpage
\appendix
\begin{center}
    \Large{\textbf{Appendix}}
\end{center}

\section{Real World Latency}
In this section, we report the latency when our mixed precision QNNs are deployed to edge devices. We test the inference speed on Raspberry Pi 3B with a 1.2 GHz 64-bit quad-core ARM Cortex-A53. We leverage the open sourced inference framework daBNN\footnote{\url{https://github.com/JDAI-CV/dabnn}} and the SIMD instruction SSHL on ARM NEON to implement Binary Decomposition. We test the speed on several specific convolutional layers in ResNet-18, and we also report the inference speed of Bi-Real Net structure. It is necessary to point out there exists no open sourced library that supports the mixed precision computation in ARM CPU. We provide a general method and there is still a lot of room for optimization of the acceleration. 
Table~\ref{tab:latency} summarizes the latency tested on ARM CPU. We can see that the binary decomposition of \textit{W1-A2} has approximately 2$\times$ latency of the binary convolution, which is similar to our theoretical estimation. Then, we test the overall inference speed of ResNet-18 under Bi-Real architecture. Note that the latency is not completely 2$\times$ because there are other overhead like img2col and data load/store.

\begin{table}[h]
    \centering
    \caption{Latency tested in different types of layers in ResNet-18. \textit{W1-A2} means weight are quantized to 1-bit and activations are quantized to 2-bit.}
    \begin{tabular}{cccccc}
    \toprule
    \multicolumn{4}{c}{\textbf{Layer Shape}} & \multicolumn{2}{c}{\textbf{Latency} (ms)} \\
    \cmidrule(l{2pt}r{2pt}){1-4}\cmidrule(l{2pt}r{2pt}){5-6}
    Kernel Size & Input Channels & Output Channels & Stride & \textit{W1-A1} & \textit{W1-A2} \\\midrule
    3 & 64 & 64 & 1 & 5.76 & 11.65\\
    3 & 128 & 128 & 1 & 5.43 & 11.46\\
    3 & 256 & 256 & 1 & 5.73 & 11.76\\
    3 & 256 & 512 & 2 & 1.65 & 3.45\\
    3 & 512 & 512 & 1 & 7.10 & 14.35\\
    \midrule
    \multicolumn{4}{l}{Bi-Real-18-ImageNet} & 277.2 & 360.8\\
    \bottomrule
    \end{tabular}
    \label{tab:latency}
\end{table}

\section{Implementation Details}
\subsection{Learning Clipping Parameter}
Recent works successfully improve the task performance of QNN by learning the clipping parameter in activations like PACT. We also adopt this strategy in learning the clipping range of activations. In particular, only one clipping parameter is maintained and learned. To specify the learning process, we split Eq.~1b to the following 3 sub-equations.
\begin{subequations}
\begin{align}
    \tilde{\bX} & = \frac{\text{clip}(\bX, 0, \alpha)}{\alpha}, \\
    \bX^b & = \text{quantize}_{b}(\tilde{\bX}),\\
    \hat{\bX} & = \alpha\times \bX^b.
\end{align}
\end{subequations}
Here, we keep Eq.~16a and Eq.~16c intact in searching the bitwidth, i.e.,
\begin{equation}
    \hat{\bX} = \alpha \times (\sum_{b_i\in\mathcal{B}_x}\frac{\exp(s_i)}{\sum_{b_j\in\mathcal{B}_x}\exp(s_j)}{\bX}^i), \text{ where }{\bX}^i=\text{quantize}_{i}(\tilde{\bX}).
\end{equation}
Then, we use the Straight-Through Estimator to approximate the gradient of $\alpha$, given by
\newcommand{\grad}[2]{\frac{\partial #1}{\partial #2}}
\begin{equation}
    \grad{\hat{\bX}}{\alpha} = \sum_{b_i\in\mathcal{B}_x}\frac{\exp(s_i)}{\sum_{b_j\in\mathcal{B}_x}\exp(s_j)}
    \bigg [{\bX}^i + \alpha \times \grad{{\bX}^i}{\alpha}\bigg ].
\end{equation}
Consequently, we have to consider two situations: $\bX>\alpha$ or $\bX \leq \alpha$. In the former case, it is simple to compute that $\tilde{\bX}=\bX^b=1$ and thus the gradients is equal to 1.
In the latter case, we can compute the gradient by:
\begin{equation}
    \grad{\hat{\bX}}{\alpha} =
    \sum_{b_i\in\mathcal{B}_x}\frac{\exp(s_i)}{\sum_{b_j\in\mathcal{B}_x}\exp(s_j)} \bigg [\hat{\bX}^i - \frac{\bX}{\alpha}\bigg ].
\end{equation}

\subsection{Implementation for Model Search}
our search space is set to $\mathcal{B}=\{1,2,3,4,5\}$. 
The strength parameters (a.k.a architecture parameters) $\mathbf{r}, \mathbf{s}$ are initialized to zero for both deterministic and stochastic optimization, as this would allocate each quantization bitwidth equal probability to be discovered.
For CIFAR10 dataset that consists of 50K training images and 10K test images, we split the training images into half for training and half for validation respectively. We first pre-train a full precision model and use it to initialize the model for searching. We use SGD with momentum of 0.9 for weights and the learning rate is set to 0.01 followed by a cosine annealing schedule. For $\mathbf{r}$, we use Adam optimizer with learning rate of 0.02 and the tradeoff parameter $\lambda$ is set to 0.06. The batch size is set to 64 for both training and validation. Weight decay for weights and $\alpha$ is set to 5e-4.
Target FLOPs is determined according to the FLOPs of 2-bit, 3-bit, and 4-bit architectures. We train the network for 60 epochs.
The temperature parameter in Eq.~\ref{eq:gumbel} is linearly decreased to 0.4 from 1.0 in the stochastic search. Precision search only takes about 6 hours for ResNet-56 on a single NVIDIA GTX 1080Ti GPU.

For ImageNet dataset which contains of 1.2M training and 50K validation images, we follow a standard data preprocessing way used in baselines~\cite{he2016resnet,jung2019qil,rastegari2016xnor,zhang2018lqnet}. Training images are randomly cropped and resized and the validation images are center-cropped to 224$\times$224. We also follow~\cite{wu2019mixed} where only 40 categories are random sampled for searching. 
Note that we split 0.8x images to training set and 0.2x images to validation set, as we find that the validation loss cannot be minimized too much if we split a large set for validation.
We set batch size to 256 for both training and validation.
The initialization follows those in CIFAR10 experiments. We also use SGD with momentum of 0.9 to jointly train the weights and the clipping parameters, following by a cosine annealing schedule. Weight decay is set to 1e-4.
For $\mathbf{r}$, we use Adam optimizer with learning rate of 0.02 and the tradeoff parameter $\lambda$ is set to 0.03.  
We do not quantize the first and the last layers as implemented in the prior works~\cite{choi2018pact,wu2019mixed,zhou2016dorefa}. We search the model for 60 epochs. Precision search only takes about 10 hours for ResNet-18 on 4 Tesla V100 GPUs.

\subsection{Implementations for Model Retraining}
During the model search, we save the strength parameters with the highest validation accuracy and directly use the $\argmax$ to select the optimal precision. For CIFAR10, we use the original 50K training images as the training set and 10K test images for the test. 
During model evaluation, there is no architecture parameter, and we do not have FLOPs penalty in the training objective. 
We use SGD with momentum of 0.9 to optimize the weights and the clipping parameters. The learning rate is set to 0.04 and followed by a cosine annealing learning rate. Weight decay (i.e., the L2 regularization term) is set to 5e-4 for high-bit models and 1e-4 for low-bit models as the low-bit QNN has less probability to overfit the training data. Batch size is set to 128 in the retraining.
We use progressive initialization, that is to say, we progressively retrain the model from highest FLOPs (precision) and use it to initialize the next model. The first model are initialized from the full precision models. The clipping parameter $\alpha$ is initialized to 6.0 in the first model. We train the model for 300 epochs. Regarding the uniform precision QNN and the random searched models in CIFAR10, we also use the same configuration in here. 

For retraining the models on ImageNet dataset, we also followed the prior data preprocessing pipeline adopted in existing works. The test images are centered cropped to 224x224 and the training images are random resized and cropped to 224x224. Batch size is set to 1024, and the weight decay is set to 1e-4 to 2e-5 depends on the bitwidth. Other training configurations like learning rate and its scheduler are the same with CIFAR10 experiments. Note that we use label refinery in order to fairly compare the performance with DNAS.  

\section{Results of ResNet-34}
We report the Top-1 accuracy and Top-5 accuracy on ResNet-34 in the following table. It can be seen that both EBS-Det and EBS-Sto consistently outperform the performance of other techniques, such as DNAS and DSQ with similar FLOPs.

\begin{table}[h]
\centering
\captionof{table}{Accuracy and computations cost comparison of ResNet-34 between EBS and existing methods in ImageNet.}
\begin{adjustbox}{max width=0.8\linewidth}
    \begin{tabular}{lcc cccc}
    \toprule
    \multirow{2}{6em}{\textbf{Methods}} & \multicolumn{2}{c}{\textbf{Precision}} &  \multicolumn{4}{c}{\textbf{ResNet-34}}\\
    \cmidrule(l{2pt}r{2pt}){2-3} \cmidrule(l{2pt}r{2pt}){4-7} 
    & Weights & Activations & Top-1 & Top-5 & FLOPs & Saving\\
    \midrule
    Full Prec. & 32-bit & 32-bit & 73.7 & 91.3 & 3.68 G & 1.0$\times$ \\
    \midrule
    BCGD & 4-bit & 4-bit & 70.8 & - & 1096 M & 3.36$\times$ \\
    DSQ & 4-bit & 4-bit & 72.8 & - & 1096 M & 3.36$\times$ \\
    EBS-Det & \flexible & \flexible & \textbf{73.5} & \textbf{91.2} & {1104 M} & {3.33$\times$} \\
    EBS-Sto & \flexible & \flexible & 73.4 & \textbf{91.2} & \textbf{1073 M} &  \textbf{3.42$\times$}\\
    \midrule
    LQ-Net & 3-bit & 3-bit & 71.9 & 90.2 & 669 M & 5.50$\times$ \\
    DSQ & 3-bit & 3-bit & 72.5 & - & 669 M & 5.50$\times$ \\
    EBS-Det & \flexible & \flexible & 73.0 & \textbf{90.8} & 654 M & \textbf{5.62}$\times$ \\
    EBS-Sto & \flexible & \flexible & \pink\textbf{73.1} & \pink\textbf{90.8} & \pink\textbf{648 M} & \pink\textbf{5.67}$\times$ \\
    \midrule
    LQ-Net & 2-bit & 2-bit & 69.8 & 89.1 & 363 M & 10.1$\times$ \\
    LQ-Net & 1-bit & 2-bit & 66.6 & 86.9 & 241 M & 15.3$\times$ \\
    DSQ & 2-bit & 2-bit & 70.0 & - & 363 M & 10.1$\times$ \\
    EBS-Det & \flexible & \flexible &  70.3 & 89.3 & 354 M & 10.4$\times$ \\
    EBS-Sto & \flexible & \flexible & \purple\textbf{70.6} & \purple\textbf{89.5} & \purple\textbf{343 M} & \purple\textbf{10.7$\times$} \\
    \midrule
    DNAS & \flexible & \flexible & 74.1 & - & 1176 M & 3.13$\times$ \\
    (+label refinery) & \flexible & \flexible & 73.2 & - & 825 M & 4.46$\times$ \\
    \midrule
    EBS-Det & \flexible & \flexible & \blue\textbf{74.3} & \blue\textbf{91.7} & \blue1104 M & \blue3.33$\times$ \\
    (+label refinery) & \flexible & \flexible & \blue\textbf{73.4} & \blue\textbf{91.1} & \blue654 M & \blue5.62$\times$ \\
    \bottomrule
    \end{tabular}
    \end{adjustbox}
\label{tab2}
\end{table}

\end{document}

%% file: math_commands.tex
\usepackage{amsmath,amsfonts,bm}

\def\eqref#1{equation~\ref{#1}}

\def\1{\bm{1}}

\newcommand{\test}{\mathcal{D_{\mathrm{test}}}}

\DeclareMathAlphabet{\mathsfit}{\encodingdefault}{\sfdefault}{m}{sl}
\SetMathAlphabet{\mathsfit}{bold}{\encodingdefault}{\sfdefault}{bx}{n}

\newcommand{\R}{\mathbb{R}}

\DeclareMathOperator*{\argmax}{arg\,max}
\DeclareMathOperator*{\argmin}{arg\,min}